\documentclass{article}
\pdfoutput=1

\usepackage[final]{neurips_2022}

\usepackage[utf8]{inputenc} 
\usepackage[T1]{fontenc}    
\usepackage[hidelinks,colorlinks=true,linkcolor=green]{hyperref}
\usepackage{url}            
\usepackage{booktabs}       
\usepackage{amsfonts}       
\usepackage{nicefrac}       
\usepackage{microtype}      
\usepackage{xcolor}         
\usepackage{enumerate}
\usepackage{graphicx}
\usepackage{multirow}
\usepackage{multicol}
\usepackage{amsmath}
\usepackage{subfig}
\usepackage{makecell}
\usepackage{enumitem}
\usepackage{floatrow}
\title{Long-Form Video-Language Pre-Training with Multimodal Temporal Contrastive Learning}

\newcommand*{\affaddr}[1]{#1} 
\newcommand*{\affmark}[1][*]{\textsuperscript{#1}}
\newcommand*\samethanks[1][\value{footnote}]{\footnotemark[#1]}

\author{%
Yuchong Sun\affmark[1]\thanks{This work was performed when Yuchong Sun and Hongwei Xue were visiting Microsoft Research as research interns.}, Hongwei Xue\affmark[2]\samethanks[1], Ruihua Song\affmark[1]\thanks{Ruihua Song and Bei Liu are the corresponding authors.}, Bei Liu\affmark[3]\samethanks[2], Huan Yang\affmark[3], Jianlong Fu\affmark[3]\\
\normalsize
\affaddr{\affmark[1]Renmin University of China, Beijing, China}, \\
\affaddr{\affmark[2]University of Science and Technology of China, Hefei, China},\\
\affaddr{\affmark[3]Microsoft Research, Beijing, China},\\
\normalsize
\affmark[1]{\tt \{ycsun, rsong\}@ruc.edu.cn}, 
\affmark[2]{\tt gh051120@mail.ustc.edu.cn}, \\
\affmark[3]{\tt \{bei.liu, huayan, jianf\}@microsoft.com}
}

\begin{document}

\maketitle

\begin{abstract}
Large-scale video-language pre-training has shown significant improvement in video-language understanding tasks.
Previous studies of video-language pre-training mainly focus on short-form videos (i.e., within 30 seconds) and sentences, leaving long-form video-language pre-training rarely explored. 
Directly learning representation from long-form videos and language may benefit many long-form video-language understanding tasks. However, it is challenging due to the difficulty of modeling long-range relationships and the heavy computational burden caused by more frames.
In this paper, we introduce a \textbf{L}ong-\textbf{F}orm \textbf{VI}deo-\textbf{LA}nguage pre-training model (LF-VILA) and train it on a large-scale long-form video and paragraph dataset constructed from an existing public dataset. 
To effectively capture the rich temporal dynamics and to better align video and language in an efficient end-to-end manner, we introduce two novel designs in our LF-VILA model.
We first propose a Multimodal Temporal Contrastive (MTC) loss to learn the temporal relation across different modalities by encouraging fine-grained alignment between long-form videos and paragraphs.
Second, we propose a Hierarchical Temporal Window Attention (HTWA) mechanism to effectively capture long-range dependency while reducing computational cost in Transformer.
We fine-tune the pre-trained LF-VILA model on seven downstream long-form video-language understanding tasks of paragraph-to-video retrieval and long-form video question-answering, and achieve new state-of-the-art performances.
Specifically, our model achieves 16.1\%  relative improvement on ActivityNet paragraph-to-video retrieval task and 2.4\% on How2QA task, respectively. We release our code, dataset, and pre-trained models at \url{https://github.com/microsoft/XPretrain}.

\end{abstract}
\section{Introduction}
\label{intro}

In recent years, research on video understanding has attracted extensive attention due to the huge amount of videos available everywhere in our daily life. 
Previous research works on video understanding~\cite{feichtenhofer2019slowfast,feichtenhofer2016two-stream-fusion,simonyan2014two-stream,tran2015c3d,zhao2017tsn} mainly focus on short-form video (i.e., $<$ 30 seconds) analysis and the semantics are limited to certain types (e.g., actions, scenes).
However, there are so many long-form videos (i.e., $>$ 30 seconds)~\cite{wu2021towards-longform} in real scenarios. Human annotated labels (e.g., actions) are difficult to cover the rich semantic and dynamic information contained in those videos.
\begin{figure}
    \centering
    \includegraphics[width=\columnwidth]{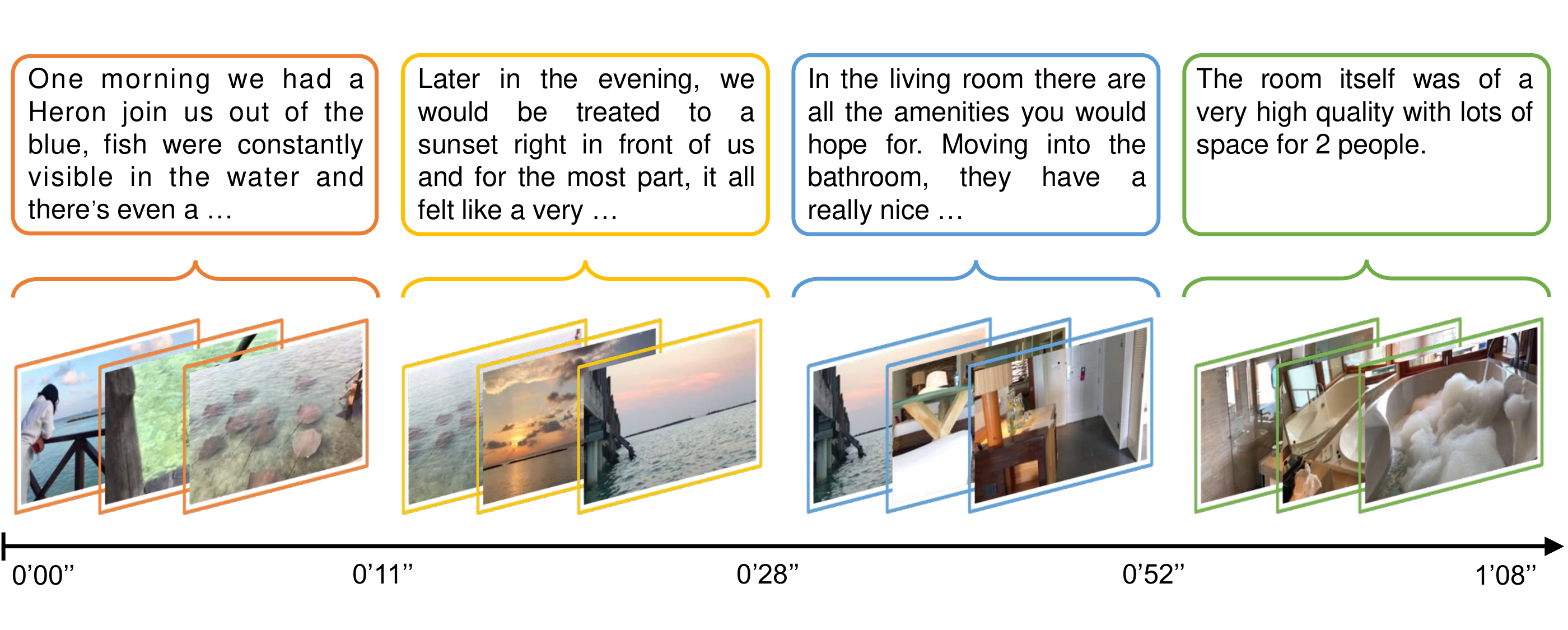}
    \caption{An example of long-form video-paragraph pair with several clips and sentences. It contains a complicated storyline and a rich temporal dynamic. Each sentence can only describe a short clip, and understanding the whole video needs the ability of long-range spatial-temporal reasoning.\vspace{-5mm}}
    \label{fig:overview}
    \vspace{-5mm}
\end{figure}
On the other hand, the video-language pre-training paradigm provides a way to learn cross-modal representation from video and language pairs and shows promising results on various high-level video understanding tasks joint with language~\cite{bain2021frozen,li2020hero, xue2021hdvila,zellers2021merlot}. However, these studies mainly focus on short-form videos. In this paper, we explore directly exploiting long-form video and language pairs for pre-training to benefit a wide range of long-form video-language understanding tasks.

Although long-form video-language joint learning has been explored in downstream tasks~\cite{ging2020coot,li2020hero,li2021value,liu2020violin,yu2019activitynetqa,zhang2018hse,zhuang2020multichannel-vqa}, they either use pre-extracted video features which lead to the sub-optimal problem, or utilize image encoder to extract frame features that fail to model the long-range dependency in long-form videos. Recent works~\cite{bain2021frozen,bertasius2021timesformer,liu2021videoswin} have shown that a video Transformer~\cite{vaswani2017transformer} backbone helps to capture long-range dependency in an end-to-end fashion.
An intuitive way for long-form video-language pre-training is to adopt a video Transformer based short-form video-language pre-training model~\cite{bain2021frozen,xue2021hdvila} with long-form data. 
However, there are two main challenges in such a design.
First, long-form videos often contain more complicated storylines and richer temporal dynamics as shown in Fig.~\ref{fig:overview}. Simply aligning video and paragraph using a vanilla contrastive loss like previous models~\cite{bain2021frozen,xue2021hdvila} will ignore the temporal relation between clips and sentences, thus hindering the quality of learned representation.
Second, feeding more frames in a Transformer based video encoder will largely increase computational cost considering the self-attention operation.

To overcome the above challenges, we propose a \textbf{L}ong-\textbf{F}orm \textbf{VI}deo-\textbf{LA}nguage pre-training model (LF-VILA) with two novel designs.
First, to better align long-form video-language pairs and learn the temporal relationship between visual and language modalities, we propose a Multimodal Temporal Contrastive (MTC) loss that learns temporal alignment between video clips and single sentences. MTC encourages similarity between two modalities to be consistent with their temporal relationship. In other words, the embedding distance between a video clip and a sentence closer in time should be smaller than its distance with sentences that are far in time. Combining with global alignment between video and paragraph, MTC ensures the model capture the temporal relation between video clips and single sentences and further helps to improve the quality of joint representation. 

Second, to utilize the advantage of the Transformer for capturing long-range dependency while efficiently processing more frames for end-to-end training, we propose a Hierarchical Temporal Window Attention (HTWA) mechanism.
As shown in Fig.~\ref{fig:overview}, the frames sparsely sampled from a long-form video have large spatial and motion gaps, thus directly computing self-attention on all frames in all layers of the Transformer is inefficient and unnecessary.
Instead, we only learn the attention between adjacent frames in the first few layers that focus more on details of spatial and temporal information.
Then we gradually expand the window size in the following layers, where the high-level representation enables the model to better capture the relation between frames far apart. The computational cost is largely reduced with the proposed HTWA mechanism.

We conduct experiments and evaluate LF-VILA on seven downstream long-form video-language understanding tasks of paragraph-to-video retrieval and long-form video question-answering. We surpass the state-of-the-art models pre-trained on short videos by a large margin. Our results demonstrate the benefit of modeling long-range dependency for long-form videos. We also verify the effectiveness of our proposed MTC loss and HTWA mechanism through ablation studies.

Our contributions are summarized as follows:
(1) We are the first to study end-to-end long-form video-language pre-training with large-scale video-paragraph data to benefit long-form video understanding. 
(2) We propose an MTC loss to capture the temporal relationship between clips and sentences while improving the joint representation of long-form video and language.
(3) We design an HTWA mechanism for the video Transformer backbone, which can capture the long-range dependency in long-form videos effectively and efficiently.
(4) We verify the effectiveness of our LF-VILA model on a wide range of downstream long-form video-language understanding tasks. Our model achieves state-of-the-art performance on four paragraph-to-video retrieval tasks and three long-form video question-answering tasks.
\section{Related Work}
\label{work}

\subsection{Video Representation}

Most previous video encoders utilize 3D-CNN based backbones~\cite{carreira2017i3d-k600,tran2015c3d,xie2018s3d}. These models show promising performance on short-form video understanding tasks, such as action classification and detection~\cite{carreira2017i3d-k600,caba2015activitynet,goyal2017something}. However, CNN has a limited receptive field and cannot effectively capture long-range dependency. Recent works have extended Vision Transformer~\cite{dosovitskiy2020vit} for video representation and demonstrated the benefit of long-range temporal learning~\cite{bertasius2021timesformer,liu2021videoswin}. To reduce the computational cost, TimeSformer~\cite{bertasius2021timesformer} introduces a factorized spacetime attention, while Video Swin-Transformer~\cite{liu2021swin} restricts self-attention in a local 3D window. However, TimeSformer~\cite{bertasius2021timesformer} is still computationally expensive when the number of input frames becomes large. Video Swin-Transformer~\cite{liu2021videoswin} adopts a fix-sized temporal window which is not suitable for videos with large duration.
We propose hierarchical temporal window attention to effectively learn the long-range dependency in long-form videos while reducing the computational cost.

\subsection{Long-form Video Understanding}
Long-form video understanding is less explored in previous studies. Some works use long-term context for improving recognition performance~\cite{shvets2019leveraging,wu2019long-feature-bank}.
Typical long-form video understanding tasks contain shot or event boundary detection~\cite{baraldi2015shot-scene} and temporal action detection~\cite{caba2015activitynet}, but these tasks cannot reveal the ability of a high-level understanding of the model.
Jointly understanding long-form videos with language is a way to discover the rich semantics contained in videos and many benchmarks are proposed recently, such as paragraph-to-video retrieval~\cite{anne2017didemo,bain2020cmovie,krishna2017actnetcaption,oncescu2021queryd} and long-form video question- answering~\cite{li2020hero,li2021value,liu2020violin,yu2019activitynetqa}.
Previous works that explore these tasks mostly use pre-extracted features, which hinder the performance because of sub-optimal features~\cite{ging2020coot,li2020hero,zhang2018hse,zhuang2020multichannel-vqa}.
We study end-to-end long-form video-language pre-training and transfer to long-form video-language understanding tasks.

\subsection{Video-Language Pre-training}
Inspired by the success of image-language pre-training~\cite{chen2020uniter,huang2021soho,huang2020pixelbert, huo2021wenlan,jia2021align,Alec2021CLIP,xue2021probing}, video-language pre-training is also explored recently. However, these works mainly focus on short-form videos~\cite{bain2021frozen,miech2020milnce,miech2019howto100m,xue2021hdvila}. Some works use 3D-CNN as a video backbone~\cite{miech2020milnce, miech2019howto100m}. To utilize the advancement of Transformer, some works use sparsely sampled frames to reduce the computation requirements~\cite{bain2021frozen,xue2021hdvila}. One key factor for learning good representation is using contrastive loss to align multi-modal features~\cite{bain2021frozen,miech2020milnce,miech2019howto100m,xue2021hdvila,zellers2021merlot}. We further design a multimodal temporal contrastive loss to conduct fine-grained alignment between long-form videos and paragraphs.
The power of the pre-training model is largely dependent on the amount of training data, some works built large-scale video-language datasets~\cite{miech2019howto100m,xue2021hdvila,zellers2021merlot}, and we build a long-from video-paragraph dataset based on HD-VILA-100M    ~\cite{xue2021hdvila}. There are several works have explored long-form video-language pre-training, HERO~\cite{li2020hero} uses pre-extracted features, while MELORT~\cite{zellers2021merlot} uses an image encoder to separately encode frames which ignores joint spatial-temporal representation. Different from them, we use a video Transformer backbone and end-to-end pre-training on large-scale long-form video-paragraph datasets.
\begin{figure}
    \centering
    \includegraphics[width=\columnwidth]{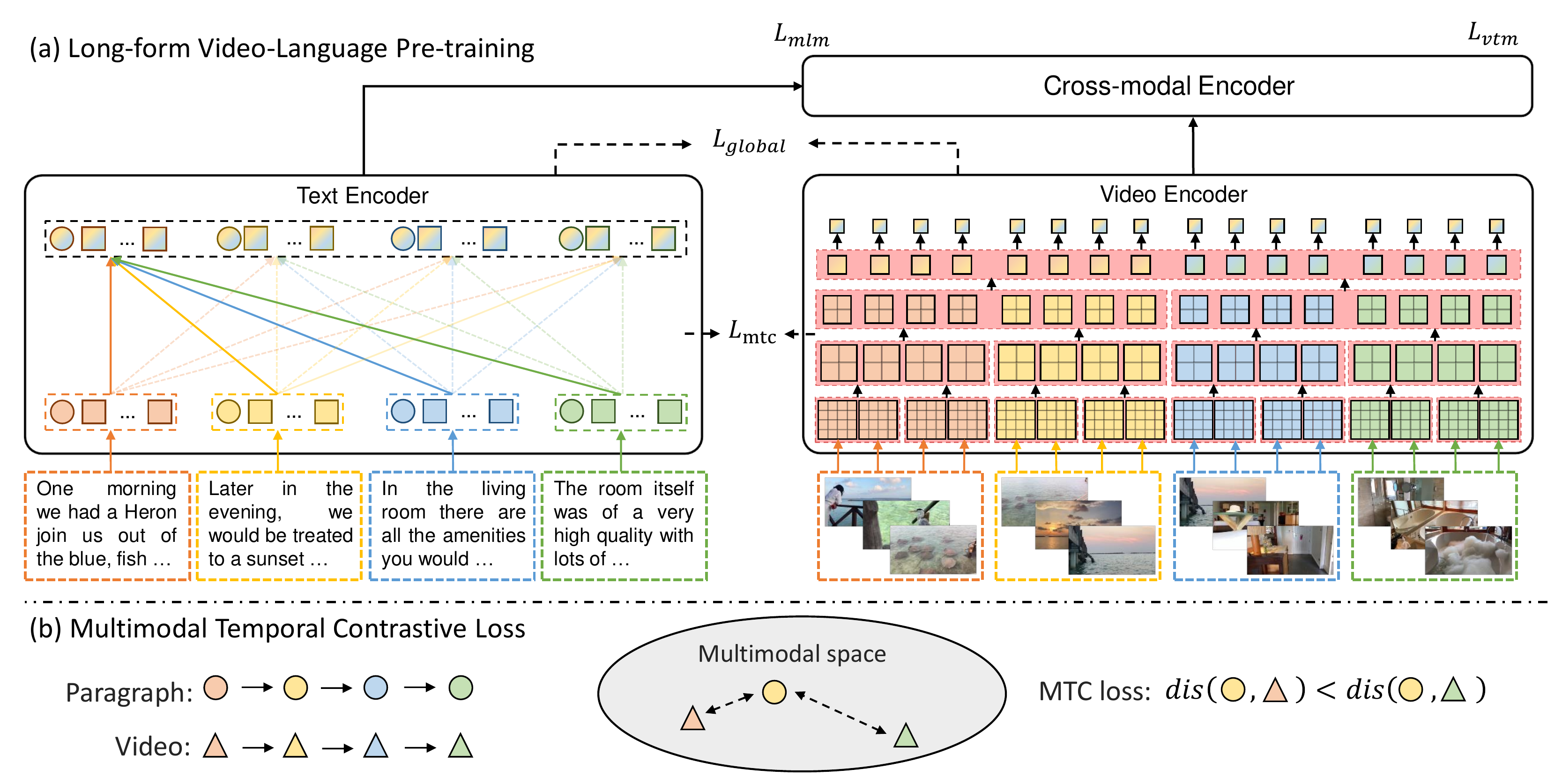}
    \caption{The framework of (a) Long-Form VIdeo-LAnguage pre-training model (LF-VILA) and illustration of (b) Multimodal Temporal Contrastive (MTC) learning. (a) LF-VILA consists of a text encoder, a video encoder, and a cross-modal encoder. In the text encoder, attention computing is first within each sentence then the whole paragraph. The pink boxes in the video encoder illustrate the proposed Hierarchical Temporal Window Attention (HTWA) mechanism. (b) the MTC loss aligns two sequences of representations (e.g., clip and sentence representations in our case), the distance of two element's representations is smaller when they are closer in time.\vspace{-5mm}}
    \label{fig:framework}
\end{figure}

\section{Approach}
\label{method}

In this section, we first show the overall architecture of the proposed Long-Form VIdeo-LAnguage pre-training model (LF-VILA) in Sec.~\ref{model_arch}. Then we explain our proposed Multimodal Temporal Contrastive (MTC) loss for learning cross-modal temporal relationship in Sec.~\ref{mtcl}, followed by the designed Hierarchical Temporal Window Attention (HTWA) mechanism for efficient video encoder in Sec.~\ref{tem_window}. Finally, we introduce the pre-training pipeline with target pre-training tasks in Sec.~\ref{pipeline}.

\subsection{Model Architecture}\label{model_arch}
As illustrated in Fig.~\ref{fig:framework}, our proposed Long-Form VIdeo-LAnguage pre-training model (LF-VILA) consists of three parts: a video encoder $E_V$, a text encoder $E_T$ and a cross-modal encoder $E_C$. With a video and a paragraph as input, we first pass them to the video encoder $E_V$ and the text encoder $E_T$ for embedding learning, respectively. Then we concatenate visual and language embeddings as input to the cross-modal encoder $E_C$, where further cross-modal joint learning is conducted. The details of these encoders are as follows.

\textbf{Text Encoder.}
The text encoder $E_T$ is based on Transformer network~\cite{vaswani2017transformer}. We divide it into two parts: sentence-level and paragraph-level encoding. In the first several layers, self-attention is conducted within word tokens from the same sentence, and sentence embedding can be learned individually. 
In higher layers, we add segment embedding to distinguish each sentence, and the attention computation is extended to all word tokens of the paragraph to output paragraph representation. 

\textbf{Video Encoder.} 
Our video encoder is also stacked by Transformer layers. In particular, we design a Hierarchical Temporal Window Attention (HTWA) mechanism for efficient attention computation. Given a long-form video that has $M$ clips, we sample $N$ frames from each clip and divide each raw frame into $H \times W$ patches. Then the $M\times N \times H\times W$ patches are encoded by $E_V$. With our designed HTWA mechanism, the temporal window is gradually expanded, so that we can get hierarchical feature maps with different temporal receptive fields. In addition, video features with the same temporal window size as clip frame number $N$ in the middle layer can be utilized as the clip representation for fine-grained alignment with sentences.

\textbf{Cross-modal Encoder.}
The cross-modal encoder $E_C$ consists of Transformer layers. Visual and language embeddings from the output of $E_V$ and $E_T$ are concatenated as the input to $E_C$. Self-attention is used to capture the joint relation between visual and language modalities. $E_C$ outputs the representation of [CLS] token and each textual and visual token.

\subsection{Multimodal Temporal Contrastive Learning}\label{mtcl}
Contrastive learning is widely used in previous multimodal pre-training works to align different modalities such as image-language and video-language. 
The goal of this loss is to pull the representation of matched pairs close to each other and push unmatched pairs away from each other.
However, when aligning long-form videos and paragraphs, the vanilla contrastive loss neglects the temporal relation between clips and sentences, which is important for capturing the complex temporal dynamics in long-form videos.
To better learn the temporal relationship between different modalities, we propose a Multimodal Temporal Contrastive (MTC) loss, which can be applied to align sequences in different modalities such as video and paragraph.  

We assume that the distance of two elements' representations in different modalities should be consistent with their temporal distance. Specifically in our model, MTC loss encourages the video clip embedding to be more similar to its neighbor sentence embedding than sentences of long distance in the same paragraph.
For example in Fig.~\ref{fig:framework}, 
given two sequences of representation $ v_i = \{v_i^1, v_i^2,..., v_i^M\}$ and $t_i = \{t_i^1, t_i^2,..., t_i^M\}$ from the $i$-th sample, we first sample an anchor set $\mathcal{A}$ of $k$ representations from $v_i$, then we sample a set of representations $\mathcal{K}$ from $t_i$. For each $v_i^p$ in $\mathcal{A}$, we treat $t_i^{q^+}$ as positive, where $|p-q^+| = \min(|p-q|), t_i^q \in \mathcal{K}$ \footnote{Since we consider $M=4$ in this work, here $\mathcal{A}$ and $\mathcal{K}$ are randomly sampled. When we extend this loss for much longer sequences, we need to restrict the maximum distance between positive pairs.}. We also randomly sample some representations from $t_j, j\neq i$ as $\mathcal{N}$ which is used to stable the training. Then the MTC loss is calculated by applying an InfoNCE loss:
\begin{equation}
\mathcal{L}_{mtc}(v_i, t_i) =-\frac{1}{k}\sum_{v_i^p \in \mathcal{A}}\log \frac{\exp \left(s(v_i^p, t_i^{q^+})\right)}{\sum_{t_i^q \in \mathcal{K}} \exp \left(s(v_i^p,t_i^q)\right) + \sum_{t_{j}^{q} \in \mathcal{N}} \exp \left(s(v_i^p,t_{j}^{q})\right)},
\end{equation}


where $s(f_1,f_2)=f^T_1 \cdot f_2 / \tau$, $\tau$ is the temperature.

We obtain clip representations of $v_i$ and sentence representations $t_i$ from the output of the first part of the video encoder and text encoder, respectively. Then the MTC loss can be obtained by:
\begin{equation}
    \mathcal{L}_{mtc}^{v2t} =-\frac{1}{B} \sum_{i=1}^{B} \mathcal{L}_{mtc}(v_i, t_i), \;
    \mathcal{L}_{mtc}^{t2v} =-\frac{1}{B} \sum_{i=1}^{B} \mathcal{L}_{mtc}(t_i, v_i),
\end{equation}
the overall MTC loss $\mathcal{L}_{mtc}$ is the average of $\mathcal{L}_{mtc}^{v2t}$ and $\mathcal{L}_{mtc}^{t2v}$.

\subsection{Hierarchical Temporal Window Attention}\label{tem_window}
Directly feeding all sampled frames of a long-form video to a vanilla Transformer network~\cite{vaswani2017transformer} for global self-attention learning will heavily increase the computational cost. The cost is quadratic with respect to the number of frames.
One possible way is to apply a small fixed temporal window. However, such a design will neglect to learn the relationship between different clips of one video, which is essential for long-form video understanding. To capture long-range dependency in long-form videos efficiently and effectively, we propose a Hierarchical Temporal Window Attention (HTWA) mechanism by gradually increasing temporal window size in Transformer layers.

A temporal window is used to restrict attention computing between frames in the temporal dimension. Given a 3-D input tensor of $T' \times H' \times W'$ patches, where $T'$ is the number of time steps and $H' \times W'$ denotes the number of spatial patches. We divide the input tensor to $M_l$ windows along $T'$ dimension, where $l$ denotes the Transformer layer. Then multi-head attention is applied within each window, and the output is the combination of all attention windows as follows:
\begin{equation}
\begin{aligned}
    & a_i = MHA\left(z_i\right), i \in [1,2,...,M_l], \\
    & a = Concat\left(a_1, a_2, ..., a_{M_l}\right),
\end{aligned}
\end{equation}
where $z_i$ is the token embeddings belonging to $i$-th window, $MHA$ is multi-head attention, $a_i$ is the attention weights of $i$-th window, and $a$ is the attention weights with the shape of $T' \times H' \times W'$.

Compared with short-form videos, long-form videos have two distinguishing characteristics that we should consider for better understanding. First, there are large motion gaps (dynamics) between frames that are sparsely sampled. Second, it is important to build long-range relationships to understand the whole video. Considering the above properties, we start to build connections between adjacent frames within small window sizes (e.g., 2) in the first few layers. 
Then the temporal window size is gradually increased to capture longer-range dependency as the semantics learned become high-level.
Specifically, we use the temporal window size equal to the number of frames in a video in the last several layers, so that the full context can be attended to learn the global relationship.

\subsection{Pre-training Pipeline}\label{pipeline}

We adopt a two-stage pre-training as previous works did~\cite{xue2021hdvila} since modal-independent design enables to learn powerful single-modality embedding for downstream tasks. In the first stage, the video encoder and text encoder are learned independently with a video-text alignment task. In the second stage, text embedding and video embedding are concatenated as input to the cross-model encoder for joint representation learning. We adopt Masked Language Modeling (MLM) and Video-Text Matching (VTM) which are widely used as pre-training tasks for learning cross-modal interaction.

\textbf{Video-Text Alignment.}
Specifically, we first train the text-encoder and video-encoder using contrastive loss to align textual and visual representations. In addition to the proposed multimodal temporal contrastive loss, we also adopt a standard contrastive loss on the global representation of long-form video and paragraph. The global contrastive loss is calculated as:
\begin{equation}
\mathcal{L}_{global}^{v2t} =-\frac{1}{B} \sum_{i=1}^{B} \log \frac{\exp \left(s(V_i, T_i)\right)}{\sum_{j=1}^{B} \exp \left(s(V_i, T_j)\right)}, \;
\mathcal{L}_{global}^{t2v} =-\frac{1}{B} \sum_{i=1}^{B} \log \frac{\exp \left(s(T_i, V_i)\right)}{\sum_{j=1}^{B} \exp \left(s(T_i, V_j)\right)},
\end{equation}
where $V_i$ and $T_i$ are the representation of $i$-th video and paragraph in the batch, respectively. The global alignment loss $\mathcal{L}_{global}$ is the average of $\mathcal{L}_{global}^{v2t}$ and $\mathcal{L}_{global}^{t2v}$.

The combination of MTC loss and global contrastive loss is used as the pre-training objective for the first stage:
\begin{equation}
\mathcal{L}_{stage1}=  \mathcal{L}_{global}+\lambda_1 \mathcal{L}_{mtc},
\end{equation}
where $\lambda_1$ denotes the weight of MTC loss compared with global contrastive loss.

\textbf{Masked Language Modeling.} 
We follow the previous vision-language pre-training works~\cite{huang2020pixelbert,li2021albef,xue2021probing} to mask word tokens and predict the ground-truth labels from the output of the cross-modal encoder, which integrates the context of other textual tokens and visual tokens:
\begin{equation}
\mathcal{L}_{mlm}=-\mathbb{E}_{(\mathcal{W}, \mathcal{V})} \log p\left(w_{i} \mid \mathcal{W}_{\backslash i}, \mathcal{V}\right),
\end{equation}
where $\mathcal{W}$ denotes the word tokens, $\mathcal{V}$ denotes the visual tokens, and $w_i$ denotes the masked token. We adopt the same masking strategy and prediction method as BERT~\cite{Devlin2018bert}.

\textbf{Video-Text Matching.}
To fuse the textual and visual information and generate a cross-modal representation, we use VTM as one pre-training task. VTM predicts whether the input paragraph and video are matched. We randomly replace the aligned video with a sampled negative with a probability of 0.5. We use a projection layer on the top of [CLS] embedding to predict two-class matching logits $y$, and then compute negative log likely-hood loss as the VTM loss:
\begin{equation}
\mathcal{L}_{vtm}=-\mathbb{E}_{(\mathcal{W}, \mathcal{V})} \log p\left(y \mid \mathcal{W}, \mathcal{V}\right).
\end{equation}

In the second stage, we freeze the video-encoder and text-encoder as~\cite{xue2021hdvila} to accelerate training. The overall loss of stage two is the combination of MLM and VTM:
\begin{equation}
\mathcal{L}_{stage2}= \mathcal{L}_{mlm}+\lambda_2 \mathcal{L}_{vtm},
\end{equation}
where $\lambda_2$ is the weight of VTM in consideration of MLM.
\begin{table}[t]
\small
    \centering
    \begin{tabular}{l l c r r r} 
    \toprule
    Dataset & Domain  & \#Video-Text Pairs & Avg. Len(sec) & Text Len & Duration(h) \\
    \midrule
    DiDeMo~\cite{anne2017didemo} & Flickr & ~~~10K & 28.0 & 29.2 &87 \\ 
    QuerYD~\cite{oncescu2021queryd} & open  & ~~~~~2K  & 278.0 & 243.8 & 200\\
    ActivityNet Captions~\cite{krishna2017actnetcaption} & action & ~~~20K & 180.0 & 48.3 & 849 \\
    Condensed Movie~\cite{bain2020cmovie} & movie  & ~~~34K  & 132.0 & 18.0  & 1.3K \\
    WebVid-2.5M~\cite{bain2021frozen} & open & ~2.5M & 18.0 & 12.0 & 13K \\
    HowTo100M~\cite{miech2019howto100m} & instruction & 136M & 3.6 & 4.0 & 135K \\
    HD-VILA-100M~\cite{xue2021hdvila} & open & 103M & 13.4 & 32.5 & 372K \\
    \midrule
    LF-VILA-8M & open & ~8.5M & 100.2 & 307.9 & 236K \\
    \bottomrule
    \end{tabular}
    \caption{Statistics of LF-VILA-8M and its comparison with existing video-language datasets.}
    \label{tab:datasets}
\end{table}

\begin{table}[t]
    \small
    \centering
    \caption{Results of paragraph-to-video retrieval on ActivityNet Captions dataset~\cite{krishna2017actnetcaption}.}
    \begin{tabular}{l c c c c c} 
    \toprule
    Method &Pre-training Dataset & R@1 $\uparrow$ & R@5 $\uparrow$ & R@50 $\uparrow$ & MedR $\downarrow$ \\
    \cmidrule{1-6}
    HSE~\cite{zhang2018hse} & - & 20.5 & 49.3 & - & -  \\
    ClipBERT~\cite{lei2021clipbert} & COCO~\cite{chen2015mscoco}, Visual Genome~\cite{krishna2017visualgenome} & 21.3 & 49.0& -& 6.0  \\
    HD-VILA~\cite{xue2021hdvila}  &HD-VILA-100M~\cite{xue2021hdvila}  & 28.5 & 57.4 & 94.0 & 4.0 \\
    Frozen~\cite{bain2021frozen} &CC3M, WebVid-2.5M~\cite{bain2021frozen} &28.8 &60.9 &- & 3.0 \\
    Support Set~\cite{patrick2020support}  &HowTo100M~\cite{miech2019howto100m} & 29.2 & 61.6 & 94.7 & 3.0 \\
    TACo~\cite{yang2021taco} &HowTo100M~\cite{miech2019howto100m} & 30.4  & 61.2 & 93.4 & 3.0 \\
    \cmidrule{1-6}
    LF-VILA (Ours)  &LF-VILA-8M & \textbf{35.3} & \textbf{65.4}&  \textbf{95.0} & \textbf{3.0} \\ 
    \bottomrule
    \end{tabular}
    \label{tab:actnetret}
\end{table}

\begin{table}[t]
    \small
    \centering
    \caption{Results of paragraph-to-video retrieval on two datasets. * denotes results by our re-implementation.}
    \subfloat[Result on DiDeMo dataset~\cite{anne2017didemo}.\label{tab:didemoret}]{
    \begin{tabular}{l c c c c} 
    \toprule
    Method  & R@1 $\uparrow$ & R@5 $\uparrow$ & R@10 $\uparrow$  \\
    \cmidrule{1-4}
    FSE~\cite{zhang2018hse} & 13.9 & 36.0 & -  \\
    ClipBERT~\cite{lei2021clipbert} & 20.4 & 48.0 & 69.0  \\
    HD-VILA~\cite{xue2021hdvila}  & 28.8 & 57.4 & 69.1  \\
    Frozen~\cite{bain2021frozen}  & 31.0 & 59.8 & 72.4  \\
     \cmidrule{1-4}
    LF-VILA (Ours)  & \textbf{35.0} & \textbf{64.5}&  \textbf{75.8} \\ \bottomrule
    \end{tabular}
    }
    \hspace{2mm}
    \subfloat[Result on QuerYD dataset~\cite{oncescu2021queryd}.\label{tab:querydret}]{
    \begin{tabular}{l c c c c} 
    \toprule
    Method  & R@1 $\uparrow$ & R@5 $\uparrow$ & R@10 $\uparrow$ \\
    \cmidrule{1-4}
    MOEE~\cite{miech2018moee} & 11.6 & 30.2 & 43.2  \\ 
    CE~\cite{liu2019ce}  & 13.9 & 37.6 & 78.3 \\
    TeachText~\cite{croitoru2021teachtext} &14.4 &37.7 &50.9\\
    Frozen*~\cite{bain2021frozen} &53.8 &75.7 & 82.7 \\
    \cmidrule{1-4}
    LF-VILA (Ours)  & \textbf{69.7} & \textbf{85.7}&  \textbf{90.3} \\
    \bottomrule
    \end{tabular}
    }
    \vspace{-5mm}
    \label{tab:didemoqueryret}
\end{table}

\begin{table}[t]
    \small
    \centering
    \caption{Results of paragraph-to-video retrieval on Condensed Movie dataset~\cite{bain2020cmovie} from official leaderboard at \url{https://competitions.codalab.org/competitions/34124}.}
    \begin{tabular}{l c c c c c c} 
    \toprule
    Method & Geometric Mean $\uparrow$ & R@1 $\uparrow$ & R@5 $\uparrow$ & R@10 $\uparrow$  \\
    \cmidrule{1-5}
    MoEE~\cite{miech2018moee}  & ~~5.88 & ~~1.94 & ~~7.84 & 13.38 \\ 
    TeachText~\cite{croitoru2021teachtext}  & 23.15 &  12.08 & 27.40 & 37.45 \\ 
    \cmidrule{1-5}
    LF-VILA (Ours)  & \textbf{26.40} & \bf 13.56&  \textbf{32.47} & \textbf{41.79} \\ \bottomrule
    \end{tabular}
    \label{tab:cmovieret}
\end{table}

\section{Experiments}
\label{exp}
In this section, we first introduce the pre-training details and then show experiments of utilizing the pre-trained model on downstream paragraph-to-video retrieval and long-form video question-answering tasks to verify the effectiveness of our proposed LF-VILA. We also transfer LF-VILA for long-form video classification tasks to show the generalization power of the pre-trained model.

\subsection{Pre-training Details}

\textbf{Pre-training Dataset.} 
To facilitate research on long-form video understanding, We build a large-scale long-form video-paragraph dataset based on HD-VILA-100M~\cite{xue2021hdvila}, which is an existing large-scale video-language dataset with diverse categories. It contains 100 million clip-sentence pairs derived from 3.3 million YouTube videos. We keep continuous clips with at least 4 clips and construct a dataset with 8.5 million long-from videos and corresponding transcripts, namely LF-VILA-8M. Table~\ref{tab:datasets} shows the statistics of LF-VILA-8M and its comparison with existing video-language datasets.
It covers \textasciitilde 60\% of clips of the HD-VILA-100M dataset.
The average duration of each video is 100.2 seconds and the average number of words in each paragraph is 
307.9. We provide additional statistics and examples in the supplementary material.

\textbf{Implementation Details.}
During pre-training, our model samples 4 consecutive clip-sentence pairs as input. We uniformly sample 8 frames from each clip and resize the frames to $192 \times 320$. we use the WordPiece tokenizer like BERT to split each sentence into tokens with a max length of 50.
For the video encoder, we use Swin-Transformer~\cite{liu2021swin} as the backbone and integrate our proposed HTWA for frame sequence. 
Temporal window sizes are set to five stages: 2, 4, 8, 16, and 32, respectively.
We use $8 \times 8$ patches and use a fixed spatial window of $3\times 5$, the output feature is down-sampled by 64 times to $3 \times 5$.  We adopt a 12-layer Transformer network for the text encoder, with 8 layers for the first part and 4 layers for the second part. We also use a 12-layer Transformer network for the cross-modal encoder. The weight of the video encoder is initialized with Swin-Transformer pre-trained on ImageNet-21K. We use the first 12 layers of BERT-Large to initialize the weight of the text encoder, and the last 12 layers to initialize the weight of the cross-modal encoder. We provide more detailed model specifications in the supplementary material.

We use an AdamW optimizer with a learning rate of 5e-5 and warm up the learning rate for 1 epoch, followed by a linear decay, we use a weight decay of 0.05. We train our model with 32 NVIDIA Tesla V100 GPUs. For stage one, we use a batch size of 512 and train for 6 epochs. For stage two, we use a batch size of 1,536 and train for another 6 epochs. We use the model from stage one for retrieval tasks since two-stream architecture is efficient for retrieval and is widely used in previous works~\cite{Alec2021CLIP,bain2021frozen,xue2021hdvila}. Pre-trained model from stage two is applied for video QA tasks. We excluded videos that overlap with downstream tasks from the training dataset using YouTube IDs.

\subsection{Paragraph-to-Video Retrieval}
We conduct retrieval tasks on four widely-used datasets for paragraph-to-video retrieval: \textbf{ActivityNet Captions}~\cite{krishna2017actnetcaption}, \textbf{DiDeMo}~\cite{anne2017didemo} , \textbf{QuerYD}~\cite{oncescu2021queryd} and \textbf{Condensed Movie}~\cite{bain2020cmovie}.
Details of each dataset and implementation are in the supplementary material.

\begin{table}[t]
\small
\centering
\caption{Results of video question-answering tasks. We gray out some results for fair comparison because of the use of large-scale video QA triplets or the huge computational cost for pre-training.}
\subfloat[ActivityNet QA~\cite{yu2019activitynetqa}.\label{tab:actnetqa}]{
\begin{tabular}{l c}
\toprule
\multicolumn{1}{c}{Method} & Acc$\uparrow$ \\
\cmidrule{1-2}
MAR-VQA~\cite{zhuang2020multichannel-vqa} & 34.6 \\
CoMVT~\cite{seo2021look-before-you-speak} & 38.8 \\
\textcolor{gray}{VQA-T~\cite{yang2021justask}} & \textcolor{gray}{38.9} \\
\textcolor{gray}{MERLOT~\cite{zellers2021merlot}}  & \textcolor{gray}{41.4} \\
\cmidrule{1-2}
LF-VILA (Ours) &  \bf 39.9 \\
\bottomrule
\end{tabular}
}
\hspace{5mm}
\subfloat[How2QA~\cite{li2020hero}.\label{tab:how2qa}]{
\begin{tabular}{l c}
\toprule
\multicolumn{1}{c}{Method} & Acc$\uparrow$ \\
\cmidrule{1-2}
CLIP~\cite{Alec2021CLIP} by \cite{li2021value} & 69.3 \\
CLIP-SF~\cite{li2021value} & 72.9 \\
ResNet-SF~\cite{li2021value} & 74.3\\
HERO~\cite{li2020hero}  & 74.3 \\
\cmidrule{1-2}
LF-VILA (Ours) & \bf 76.1 \\
\bottomrule
\end{tabular}
}
\hspace{5mm}
\subfloat[VIOLIN~\cite{liu2020violin}. \label{tab:violin}]{
\begin{tabular}{l c }
\toprule
\multicolumn{1}{c}{Method} & Acc$\uparrow$ \\
\cmidrule{1-2}
LXMERT~\cite{tan2019lxmert} &66.3\\
Det-BERT~\cite{liu2020violin} &67.8 \\
GVE~\cite{chen2021GVE}  & 68.4 \\
HERO~\cite{li2020hero}  & 68.6 \\
\cmidrule{1-2}
LF-VILA (Ours) & \bf 70.9 \\
\bottomrule
\end{tabular}
}

\vspace{-5mm}
\label{tab:video_qa}
\end{table}

\textbf{Results.}
Tab.~\ref{tab:actnetret}\textasciitilde \ \ref{tab:cmovieret} show the results of LF-VILA on four paragraph-to-video retrieval datasets.
For the most widely used \textbf{ActivityNet Captions}~\cite{krishna2017actnetcaption} dataset, we surpass the SOTA model TACo~\cite{yang2021taco} by \textbf{16.1\%} on R@1. TACo is pre-trained on HowTo100M~\cite{miech2019howto100m} using pre-extracted feature. This demonstrates the benefit of end-to-end training and utilization of long-form video datasets. 
Compared to HD-VILA~\cite{xue2021hdvila} which is trained on 100M short-form video and sentence pairs, we use long-form videos with fewer data (\textasciitilde 60\% clips of HD-VILA-100M) and achieve \textbf{23.9\%} improvement in terms of R@1. This shows the effectiveness of our model LF-VILA in learning better alignment for long-form video and language. Compared to Frozen~\cite{bain2021frozen} which is pre-trained on a human-annotated dataset, we achieve \textbf{22.6\%} improvement on R@1 using relatively noisy but easy-to-get data.
For \textbf{DiDeMo}~\cite{anne2017didemo}, we also observe a significant improvement. In particular, we obtain \textbf{12.9\%} improvement in terms of R@1 over the SOTA model Frozen~\cite{bain2021frozen}.
On \textbf{QuerYD}~\cite{oncescu2021queryd} and \textbf{Condensed Movie}~\cite{bain2020cmovie}, we outperform the previous best models Frozen~\cite{bain2021frozen} and TeachText~\cite{croitoru2021teachtext} with a relative \textbf{29.6\%} and \textbf{12.3\%} improvement, respectively. These two datasets are challenging due to their long videos. These results show the value of long-form video-language pre-training and our LF-VILA model can better understand the storyline and temporal relations in long videos.

\subsection{Long-Form Video Question-Answering}
We conduct video question-answering tasks on three widely-used datasets for long-form video understanding: \textbf{ActivityNet-QA}~\cite{yu2019activitynetqa}, \textbf{How2QA}~\cite{li2020hero} and \textbf{VIOLIN}~\cite{liu2020violin}. Details of each dataset and implementation are in the supplementary material.

\textbf{Results.}
Tab.~\ref{tab:video_qa} shows the results of three long-form video question-answering tasks.
For \textbf{ActivityNet QA}~\cite{yu2019activitynetqa}, we outperform most previous works except MELORT~\cite{zellers2021merlot}. Note that VQA-T~\cite{yang2021justask} and MERLOT~\cite{zellers2021merlot} are specifically designed for video QA. VQA-T~~\cite{yang2021justask} uses automatically generates 69M video question-answer triplets from narrated videos for training. MERLOT~\cite{zellers2021merlot} utilizes 180M pairs of data and it needs excessive computational cost for training (\textasciitilde 30K TPU hours), while we only need \textasciitilde 5K GPU hours.
For \textbf{How2QA}~\cite{li2020hero} and \textbf{VIOLIN}~\cite{liu2020violin}, we achieve the new SOTA. This illustrates the reasoning capability of our model on long-form videos by better capturing the long-range dependency between video clips and paragraphs.

\begin{table}[t]
    \small
    \centering
    \caption{Results of Procedural Activities Classification on the COIN~\cite{tang2019coin} dataset. We gray out some results for fair comparison because of the huge computational cost for pre-training.}
    \begin{tabular}{l c c c c} 
    \toprule
    Model & Pre-training Dataset & \#Training Samples &Domain & Acc$\uparrow$ \\
    \cmidrule{1-5}
    ClipBERT~\cite{lei2021clipbert} & COCO~\cite{chen2015mscoco}, Visual Genome~\cite{krishna2017visualgenome} & 5.6M  & Open & 65.4  \\
    MIL-NCE~\cite{miech2020milnce} &HowTo100M~\cite{miech2019howto100m} & 100M & Instructional & 70.2 \\
    VideoCLIP~\cite{xu2021videoclip} &HowTo100M~\cite{miech2019howto100m} & 100M & Instructional & 72.5  \\
    SlowFast~\cite{feichtenhofer2019slowfast} &Kinetics~\cite{carreira2017i3d-k600} &370K & Action & 71.6  \\
    TimeSformer~\cite{bertasius2021timesformer} &Kinetics~\cite{carreira2017i3d-k600} &370K & Action & 83.5 \\
    TimeSformer~\cite{bertasius2021timesformer} &HowTo100M~\cite{miech2019howto100m} & 100M  & Instructional & 85.3 \\
    \textcolor{gray}{TimeSformer~\cite{bertasius2021timesformer}} & \textcolor{gray}{HowTo100M~\cite{miech2019howto100m}, wikiHow~\cite{lin2022wikihow}} & \textcolor{gray}{100M}  & \textcolor{gray}{Instructional} & \textcolor{gray}{88.9}\\
    \cmidrule{1-5}
    LF-VILA (Ours)  &LF-VILA-8M & 8M & Open & \bf 85.7 \\
    \bottomrule
    \end{tabular}
    \label{tab:coin}
\end{table}

\begin{table}[t]
    \small
    \centering
    \caption{Results of Content Understanding on the LVU~\cite{wu2021towards-longform} dataset.}
    \begin{tabular}{l c c c} 
    \toprule
    Model & Relation (Acc)$\uparrow$ & Way of Speaking (Acc)$\uparrow$ & Scene (Acc)$\uparrow$ \\
    \cmidrule{1-4}
    R101-SlowFast+NL~\cite{feichtenhofer2019slowfast,he2016resnet,wang2018non-local} & 52.4 & 35.8 & 54.7  \\
    VideoBERT~\cite{sun2019videobert} & 52.8 & 37.9 & 54.9  \\
    Object Transformer~\cite{wu2021towards-longform} & 53.1 & 39.4 & 56.9 \\
    \cmidrule{1-4}
    LF-VILA (Ours)  & \bf 61.5 & \bf 41.3 & \bf 68.0   \\
    \bottomrule
    \end{tabular}
    \label{tab:lvu}
\end{table}

\subsection{Transfering to Long-form Video Classification}
To demonstrate the improvement of LF-VILA for long-form video representation and broaden the tasks and domains of the evaluation, we evaluate LF-VILA on \textbf{COIN}~\cite{tang2019coin} and \textbf{LVU}~\cite{wu2021towards-longform}. Details of each dataset and implementation are in the supplementary material.

Tab.~\ref{tab:coin} shows the result of Procedural Activities Classification on \textbf{COIN}~\cite{tang2019coin} dataset. Our model achieves strong performance on this task, although we use out-domain videos for pre-training and our computational cost is smaller than the SOTA method (\textasciitilde 2.1K vs. \textasciitilde 7K GPU hours). We largely surpass other video-language pre-training models (e.g., ClipBERT~\cite{lei2021clipbert}, MIL-NCE~\cite{miech2020milnce} and VideoCLIP~\cite{xu2021videoclip}). 
Tab.~\ref{tab:lvu} shows the result of video classification on \textbf{LVU}~\cite{wu2021towards-longform} dataset. Our model surpasses the previous SOTA methods largely, especially on Scene and Relation classification. Especially, we outperform the video-language pre-training model VideoBERT~\cite{sun2019videobert} largely. 
The strong performances on these two benchmarks show the generalization power of our pre-trained video encoder.

\subsection{Ablation Studies}
To validate the effectiveness of LF-VILA with MTC loss and HTWA mechanism, we conduct ablation studies with a subset of data to save resources. We randomly sample 1M video-paragraph pairs from the whole LF-VILA-8M for pre-training.

\textbf{(1) Using more frames is better for long-form video-language understanding.}
Using the same backbone, we compare our model with pre-training with one clip-sentence pair for each sample as in previous models~\cite{bain2021frozen,bain2021frozen}. As shown in Tab.~\ref{tab:abl_longform}, when only using 8 frames, the performance is poor. After increasing the number of frames to 32, there is a large improvement. This also indicates that an efficient backbone to support more frames is essential for long-form video representation.

\textbf{(2) Pre-training on long-form video-language data further improves the performance significantly.}
When we use 4 continuous clips for pre-training, there is also a significant performance gain as shown in Tab.~\ref{tab:abl_longform}. The gain is larger on QuerYD dataset which consists of longer videos than ActivityNet dataset (278s vs 180s on average).

\textbf{(3) MTC loss contributes to the performance.}
As shown in Tab.~\ref{tab:abl_longform}, when combined with the MTC loss, the performance is further improved by 1.7\% on ActivityNet dataset and 1.9\% on QuerYD dataset in terms of R@1. 

\textbf{(4) HTWA mechanism achieves a better computational cost-performance tradeoff.}
In Tab.~\ref{tab:abl_window}, we compare our methods with video backbone using fixed window long-range dependency. When we apply a small fixed window size (e.g.,4), the performance is relatively poor. This indicates the limitation of a small attention window for modeling long-range dependency. When we increase the window size to 32 to cover a whole video, there is almost no improvement in performance, while the computational cost and training time increases significantly. 

\begin{table}[t]
    \small
    \centering
    \caption{Analysis of the effectiveness of long-form pre-training.}
    \begin{tabular}{c c c c c c c c c} 
    \toprule
    \multicolumn{3}{c}{Pretain}&\multicolumn{3}{c}{ActivityNet~\cite{krishna2017actnetcaption}}&\multicolumn{3}{c}{QuerYD~\cite{oncescu2021queryd}} \\ 
    \cmidrule{1-9}
    Loss & \#clips & \#frames in total & R@1$\uparrow$ & R@5$\uparrow$ & R@50$\uparrow$ &  R@1$\uparrow$ & R@5$\uparrow$ & R@10$\uparrow$ \\
    \cmidrule{1-9}
    w/o Pre-training &- &- &15.0 &40.2 & 85.8 &19.8 &46.4 &58.4\\
    \cmidrule{1-9}
    \multirow{4}{*}{$\mathcal{L}_{global}$} 
    & 1 &8 &19.3 &45.5 &87.1 &38.4 &67.5 &78.2 \\ 
    & 4 &8 &20.3 &47.0 &87.7 &41.6 &73.5 &82.1 \\
    & 1 &32 &24.3 &53.5 &92.1 &50.8 &75.0 &83.4\\
    & 4 &32 & 26.1 & 56.7 & 92.7 &55.4 &79.3 &85.5\\
    \cmidrule{1-9}
    $\mathcal{L}_{global} + \mathcal{L}_{mtc}$ & 4 &32 & \bf 27.8 & \bf 58.3 & \bf 92.8 &\bf57.3 &\bf80.9 &\bf85.8 \\
    \cmidrule{1-9}
    \end{tabular}
    \label{tab:abl_longform}
\end{table}

\begin{table}[h!]
    \small
    \centering
    \caption{Analysis of temporal window (TW).}
    \begin{tabular}{l  c  c c c c c c c} 
    \toprule
    \multirow{2}{*}{Method}&\multirow{2}{*}{TW\#}&\multirow{2}{*}{Time}&\multicolumn{3}{c}{ActivityNet~\cite{krishna2017actnetcaption}} & \multicolumn{3}{c}{QuerYD~\cite{oncescu2021queryd}} \\
     \cmidrule{4-9}
    & & & R@1$\uparrow$ & R@5$\uparrow$ & R@50$\uparrow$ &R@1$\uparrow$ & R@5$\uparrow$ & R@10$\uparrow$ \\
    \cmidrule{1-9}
    \multirow{4}{*}{Fixed} 
    &4 &0.91$\times$ &25.1 &54.9 & 91.8 &52.8 &76.3 &84.5 \\
    &8 &0.96$\times$ &25.4 & 54.8 & 91.9 &54.5 &78.1 &84.4 \\
    &16 &1.10$\times$ &26.0 & 56.1 & 91.5 &53.7 &78.6 &84.3 \\
    &32 &1.49$\times$ &26.2 & 56.4 & 92.3 &55.6 &79.3 &85.7 \\
    \cmidrule{1-9}
    HTWA  &[2-32] &1.00$\times$ & 26.1 & 56.7 &92.7 &55.4 &79.3 &85.5\\ 

    \bottomrule
    \end{tabular}
    \label{tab:abl_window}
\end{table}

\section{Conclusion}
\label{conclusion}
In this paper, we study video-language pre-training on a large-scale long-form video-paragraph dataset.
To better align long-from videos and paragraphs, we propose a Multimodal Temporal Contrastive (MTC) loss to capture the rich temporal relation between different modalities.
In addition, we design a Hierarchical Temporal Window Attention (HTWA) mechanism to be applied with an image Transformer. Our proposed Long-Form VIdeo-LAnguage pre-training model (LF-VILA) combined with MTC and HTWA can learn effective multi-modal representation by capturing long-range dependency from long-form videos efficiently.
Experiments on seven long-form video-language understanding tasks verify the effectiveness of our model.

\begin{ack}
Funding in direct support of this work: Fundamental Research Funds for the Central Universities and the Research Funds of Renmin University of China (21XNLG28). Additional revenues related to this work: Internship at Microsoft Research Asia. This work is also partially sponsored by Kuaishou  Research Collaboration Initiative.
\end{ack}

\bibliographystyle{plain}
\bibliography{main}

\section*{Checklist}


\begin{enumerate}

\item For all authors...
\begin{enumerate}
  \item Do the main claims made in the abstract and introduction accurately reflect the paper's contributions and scope?
    \answerYes{}
  \item Did you describe the limitations of your work?
     \answerYes{}
  \item Did you discuss any potential negative societal impacts of your work?
     \answerYes{}
  \item Have you read the ethics review guidelines and ensured that your paper conforms to them?
     \answerYes{}
\end{enumerate}

\item If you are including theoretical results...
\begin{enumerate}
  \item Did you state the full set of assumptions of all theoretical results?
    \answerNA{}
  \item Did you include complete proofs of all theoretical results?
    \answerNA{}
\end{enumerate}

\item If you ran experiments...
\begin{enumerate}
  \item Did you include the code, data, and instructions needed to reproduce the main experimental results (either in the supplemental material or as a URL)?
    \answerNo{}{}
  \item Did you specify all the training details (e.g., data splits, hyperparameters, how they were chosen)?
    \answerYes{}
        \item Did you report error bars (e.g., with respect to the random seed after running experiments multiple times)?
    \answerNo{}
        \item Did you include the total amount of compute and the type of resources used (e.g., type of GPUs, internal cluster, or cloud provider)?
    \answerYes{}
\end{enumerate}

\item If you are using existing assets (e.g., code, data, models) or curating/releasing new assets...
\begin{enumerate}
  \item If your work uses existing assets, did you cite the creators?
    \answerYes{}
  \item Did you mention the license of the assets?
    \answerNo{}
  \item Did you include any new assets either in the supplemental material or as a URL?
    \answerNo{}
  \item Did you discuss whether and how consent was obtained from people whose data you're using/curating?
    \answerNo{}
  \item Did you discuss whether the data you are using/curating contains personally identifiable information or offensive content?
    \answerYes{}
\end{enumerate}

\item If you used crowdsourcing or conducted research with human subjects...
\begin{enumerate}
  \item Did you include the full text of instructions given to participants and screenshots, if applicable?
    \answerNA{}
  \item Did you describe any potential participant risks, with links to Institutional Review Board (IRB) approvals, if applicable?
    \answerNA{}
  \item Did you include the estimated hourly wage paid to participants and the total amount spent on participant compensation?
    \answerNA{}
\end{enumerate}

\end{enumerate}


\appendix


This supplementary material is organized as:

\begin{enumerate}
    \item Model architecture details (Sec. ~\ref{sec:app_model})
    \item Additional pre-training details (Sec.~\ref{sec:app_training})
    \item Downstream task details (Sec.~\ref{sec:app_tasks})
    \item LF-VILA-8M dataset details (Sec.~\ref{sec:app_dataset})
    \item Limitation and Broader Impact (Sec.~\ref{sec:limitation})
    \item Responsible AI Considerations (Sec.~\ref{sec:rai}).
    
\end{enumerate}

\section{Model Architecture Details}\label{sec:app_model}
\subsection{Text Encoder Specifications}
Our text encoder $E_T$ is a 12-layer Transformer~\cite{vaswani2017transformer} encoder, with a hidden size of 1024 and attention heads of 16. 
The parameter is initialized using the first 12 layers of BERT-Large~\cite{Devlin2018bert} from HuggingFace~\footnote{https://huggingface.co/} transformers library implementation.
$E_T$ is divided into two parts, with 8 layers for the first part and 4 layers for the second part.
$E_T$ takes a paragraph that has 4 sentences as input.
We first split each sentence into tokens beginning with a [CLS] token using WordPiece tokenizer with a max length of 50. 
We add position embeddings to each sentence separately, and then encode each sentence using the first part of $E_T$ where attention computation is restricted to tokens belonging to the same sentence.
We obtain representation for each sentence on the top of the [CLS] token which is used for MTC loss.
At the beginning of the second part, we add segment embeddings to distinguish each sentence. We average the 4 [CLS] embeddings as a global [CLS] token and append it at the beginning of the  token sequence. 
Then the attention is computed among all tokens in the second part, where the number of tokens is $1+4 \times 50$. 
At the top of the global [CLS] token, we obtain paragraph representation used for global alignment.

\subsection{Video Encoder Specifications}
We apply the HTWA mechanism to an image Transformer, we choose Swin Transformer-base~\cite{liu2021swin} here. The original Swin Transformer base is divided into 4 stages, we divide stage3 into 2 parts applied with different temporal windows. Tab.~\ref{tab:video_encoder_detail} shows the specific configuration, where``downs'' means the times of spatial feature map downsampling,``siz'' means the size of the feature map, ``concat $n \times n$'' means merging $n \times n$ neighboring spatial patches,``n-d'' means the dimension of output feature,``win.sz'' means a window attention module, the original Swin Transformer uses a fixed spatial window size of $7 \times 7$, and we use an additional temporal window with gradually expanding size.
We initialize the video backbone using Swin Transformer-base parameters. There are two main modifications: (1) the patch projection layer, we divide the frame to $8 \times 8$ patches, while Swin Transformer uses $4 \times 4$ patches, so we duplicate projection weight 4 times. (2) the relative position embedding parameter, we first interpolate the original parameter to fit our spatial window size, then we duplicate 2d relative position embedding along the temporal dimension as ~\cite{liu2021videoswin}.
The temporal window size is 8 in stage3, which means the attention computation is within a clip. We downsample the feature map of the stage3 two times and then average the tokens from the same clip to obtain clip representation which is used for MTC loss. After stage5, we average all tokens to obtain video representation used for global alignment.

\begin{table}[t]
\setlength\tabcolsep{5pt}
\caption{Video Encoder Configuration.}
\label{tab:video_encoder_detail}
\centering
\begin{tabular}{c c c c c c c} 
    \toprule
    \multicolumn{3}{c}{Swin Transformer-base} & & \multicolumn{3}{c}{Video Encoder of LF-VILA} \\
    \cmidrule{5-7}
    \cmidrule{1-3}
    stage & \makecell[c]{downsp \\ (size)}  & layer & 
    &stage & \makecell[c]{downsp \\ (size)}  & layer \\ 
    \cmidrule{1-7}

    stage1 &\makecell[c]{4$\times$\\(56$\times$56)} &\makecell[c]{concat $4\times4$,\\ 128-d, \\ $[\rm win.sz. \, 7\times7]\times2$}
    &
    &stage1 &\makecell[c]{8$\times$\\(32$\times$24$\times$40)}
    &\makecell[c]{concat $8\times8$,\\ 128-d, \\ $[\rm win.sz.\, 2\times 3\times5]\times2$} \\
    \cmidrule{1-7}
    
    stage2 &\makecell[c]{8$\times$\\(28$\times$28)} &\makecell[c]{concat $2\times2$,\\ 256-d, \\ $[\rm win.sz. 7\times7]\times2$} 
    &
    &stage2 &\makecell[c]{16$\times$\\(32$\times$12$\times$20)}
    &\makecell[c]{concat $2\times2$,\\ 256-d, \\ $[\rm win.sz.\, 4\times3\times5]\times2$} \\
    \cmidrule{1-7}
    
    \multirow{3}{*}{stage3} &\multirow{3}{*}{\makecell[c]{16$\times$\\(14$\times$14)}} &\multirow{3}{*}{\makecell[c]{concat $2\times2$,\\ 512-d, \\ $[\rm win.sz. 7\times7]\times18$}}
    &
    &stage3 &\makecell[c]{32$\times$\\(32$\times$6$\times$10)}
    &\makecell[c]{concat $2\times2$,\\ 512-d, \\ $[\rm win.sz.\, 8\times 3\times5]\times14$} \\
    \cmidrule{5-7}
    &&& 
    &stage4 &\makecell[c]{32$\times$\\(32$\times$6$\times$10)}
    &\makecell[c]{512-d,\\ $[\rm win.sz.\, 16\times3 \times5]\times4$} \\

    \cmidrule{1-7}
    
    stage4 &\makecell[c]{32$\times$\\(7$\times$7)} &\makecell[c]{concat $2\times2$, \\1024-d, \\ $[\rm win.sz. 7\times7]\times2$}  
    &
    &stage5 &\makecell[c]{64$\times$\\(32$\times$3$\times$5)}
    &\makecell[c]{concat $2\times2$,\\ 1024-d, \\ $[\rm win.sz.\, 32\times3\times5]\times2$} \\
    
    \bottomrule
\end{tabular}
\end{table}

\subsection{Cross-modal Encoder Specifications}

The cross-modal encoder is also a 12-layer Transformer encoder, with a hidden size of 1024 and attention heads of 16. The parameter is initialized using the last 12 layers of BERT-Large~\cite{Devlin2018bert} from HuggingFace transformers library implementation.
We use $2 \times 3$ maxpool with stride $1 \times 1$ to downsample video feature maps and obtain 6 tokens for each frame, the number of video tokens is $32\times 6$.
We concatenate all the textual tokens and video tokens. Finally the tokens of cross-modal encoder is $1 + 50 \times 4 + 32\times 6$.

\subsection{Video Encoder Computational Cost}
We did not adopt the same backbone as HD-VILA~\cite{xue2021hdvila} or Frozen~\cite{bain2021frozen} on the new dataset because their video encoders cannot be fed with so many frames in long-form videos of LF-VILA-8M  while using more frames is critical for long-form video-language understanding. In Tab.~\ref{tab:backbone}, we measure the computational cost of our video encoder compared with HD-VILA and Frozen for encoding 8 or 32 frames.

\begin{table}[t]
    \small
    \centering
    \caption{Comparison of video encoder computational cost.}
    \begin{tabular}{l c c c c} 
    \toprule
    Model & \#Frames  & Flops & Memory \\
    \cmidrule{1-4}
    Frozen~\cite{bain2021frozen} &8  & 356 G & 4.8G  \\
    HD-VILA~\cite{xue2021hdvila} &8  & 516 G & 6.3G \\
    Frozen~\cite{bain2021frozen} &32  & 1424 G & 11.6G  \\
    HD-VILA~\cite{xue2021hdvila} &32  & 1750 G & 13.1G \\
    \cmidrule{1-4}
    LF-VILA (Ours)  &32  & \bf298G & \bf5.2G  \\
    \bottomrule
    \end{tabular}
    \label{tab:backbone}
\end{table}

\section{Additional Pre-training Details}\label{sec:app_training}
\subsection{Hyperparameters}
For the temperature parameter $\tau$ in $\mathcal{L}_{global}$ and  $\mathcal{L}_{mtc}$, we set it to 0.05 as ~\cite{bain2021frozen,xue2021hdvila}. The loss weight $\lambda_1$ in Equation 5 is set to 1.0 and the loss weight $\lambda_2$ in Equation 8 is set to 10.0 empirically. For the hyperparameters in $\mathcal{L}_{mtc}$, we set the size of $\mathcal{A}$ and $\mathcal{K}$ to 2 and the size of $\mathcal{N}$ to 3 empirically.

\subsection{More Details of LF-VILA and other Pre-training Models.}

\begin{table}[t]
    \small
    \centering
    \caption{More details of LF-VILA and other pre-training models. * means using distilled text encoder.}
    \begin{tabular}{l c c c} 
    \toprule
    Model &Frozen~\cite{bain2021frozen} & HD-VILA~\cite{xue2021hdvila} & LF-VILA (Ours)\\
    \cmidrule{1-4}
    Pre-training Dataset & \makecell[c]{CC3M~\cite{sharma2018conceptual},\\ COCO~\cite{chen2015mscoco},\\ WebVid2.5M~\cite{bain2021frozen}} &HD-VILA-100M~\cite{xue2021hdvila} & LF-VILA-8M \\
    \cmidrule{1-4}
    \#Training Examples & 6.1M & 100M & 8M\\
    \cmidrule{1-4}
    Pre-training Cost (GPU Hours) & \textasciitilde1.3K & \textasciitilde65K & \textasciitilde2.1K\\
    \cmidrule{1-4}
    \#Param & 223M(181M*) & 310M & 277M \\
    \cmidrule{1-4}
    Input Resolution & 224$\times$224 & \makecell[c]{640$\times$1024 (1 frame), \\ 160$\times$256} & 192$\times$320 \\
    \bottomrule
    \end{tabular}
    \label{tab:cost}
\end{table}
We provide more details about the pre-training cost, pre-training data, parameters and input resolution of our model and other large-scale end-to-end pre-training models in Tab.~\ref{tab:cost}.
Compared to Frozen~\cite{bain2021frozen} which is pre-trained on a human-annotated dataset, we achieve much better performance using relatively noisy but easy-to-get data. Compared to HD-VILA~\cite{xue2021hdvila}, we greatly reduce the training cost with smaller model size. LF-VILA-8M is a subset of HD-VILA-100M~\cite{xue2021hdvila} (\textasciitilde60\% clips). Thus we achieve better performance with less pre-training data, fewer parameters and lower input resolution.

\section{Downstream Task Details}\label{sec:app_tasks}

\subsection{Paragraph-to-Video Retrieval}
\textbf{Datasets.}
We conduct retrieval tasks on four paragraph-video retrieval datasets as they are widely used in previous video-language pre-training works.
\textbf{(1) ActivityNet Captions}~\cite{krishna2017actnetcaption} consist of 20K videos collected from YouTube and 100K manually annotated sentences. We follow~\cite{lei2021clipbert,xue2021hdvila,zhang2018hse} to concatenate all sentences of a video and evaluate paragraph-to-video retrieval. We fine-tune LF-VILA on the training set with 10K videos and report the result on the val1 set with 4.9K videos.
\textbf{(2) DiDeMo}~\cite{anne2017didemo} consists of 10K Flickr videos and 40K manually annotated sentences. We conduct paragraph-to-video retrieval on this dataset following~\cite{bain2021frozen,lei2021clipbert,xue2021hdvila}. We use a standard split to fine-tune LF-VILA on the training set and report the result on the test set.
\textbf{(3) QuerYD}~\cite{oncescu2021queryd} contains videos sourced from YouTube. There are 1815 videos in the training split, 388 and 390 videos for validation and testing, respectively. The dataset has 31,441 high-quality descriptions. We follow~\cite{croitoru2021teachtext,oncescu2021queryd} to evaluate paragraph-level video-retrieval~\footnote{We try our best to collect videos from YouTube, however, some videos are no longer available. We have downloaded 1628 training videos, 341 validation videos, and 346 testing videos.}.
\textbf{(4) Condensed Movie}~\cite{bain2020cmovie} consists of 35K key scenes from over 3.7K movies, each key scene is accompanied by a high-level semantic description. We follow the challenge instruction, fine-tune LF-VILA on 30K videos, and report results on the test set with 1K videos from the leaderboard.

\textbf{Implementation Details.}
We adopt the pre-training model from stage one for fine-tuning because the two-stream architecture is suitable for retrieval tasks and is widely-adopted~\cite{bain2021frozen,xue2021hdvila}. We only use $\mathcal{L}_{global}$ for fine-tuning to keep a fair comparison with previous works~\cite{bain2021frozen,xue2021hdvila}. We sample 32 frames from the videos and resize the frames to 192 × 320. We merge adjacent short sentences in a paragraph to make the number of sentences 4. We use an AdamW optimizer with an initial learning rate of 5e-6 followed by a multi-step learning rate decay. We fine-tune the model with 8 NVIDIA Tesla V100 GPUs and use a batch size of 128.

\subsection{Long-Form Video Question-Answering}
\textbf{Datasets.}
We conduct video question-answering tasks on three widely-used datasets for long-form video understanding.
\textbf{(1) ActivityNet QA}~\cite{yu2019activitynetqa} is an open-ended question-answering dataset with 25K questions from 5.8K videos. The average video length is 180 seconds. It is a benchmark to test long-term spatial-temporal reasoning. We use the official dataset split and report the result on the test set.
\textbf{(2) How2QA}~\cite{li2020hero} is a multiple-choice video question-answering dataset. It consists of 44K QA pairs from 22K 60-second clips selected from HowTo100M~\cite{miech2019howto100m} dataset, each question has one correct answer and 3 wrong answers. We use the split of ~\cite{li2021value} and report the result on the validation set.
\textbf{(3) VIOLIN}~\cite{liu2020violin} is a video-language inference task. It aims to predict whether a video entails a hypothesis or contradicts the hypothesis when given a premise video with aligned subtitles and a hypothesis sentence. It consists of 95.3K video-hypothesis pairs from 15.9K video clips, and the average length is 40 seconds. We use the official dataset split and report the result on the test-public set.

\textbf{Implementation Details.}
We fine-tune the pre-training model from stage two on downstream video question-answering tasks following \cite{xue2021hdvila}.
We uniformly sample 32 frames from each video.
For How2QA~\cite{li2020hero}, we concatenate the question with each answer from candidates, as well as subtitles. On top of the global [CLS] token of the question, we train an MLP to predict the confidence of each answer to be correct with a cross-entropy loss.
For ActivityNet QA~\cite{yu2019activitynetqa}, we encode the answers in a one-hot fashion and train an MLP classifier on top of the global [CLS] token of the question overall answer candidates with a cross-entropy loss.
For VIOLIN~\cite{liu2020violin}, we concatenate the hypothesis and subtitles, and on top of the global [CLS] token, we predict the confidence of the premise entails the hypothesis with a binary cross-entropy loss. We fine-tune the model with 8 NVIDIA Tesla V100 GPUs. 

\subsection{Long-form Video Classification}
\textbf{Datasets.}
We conduct video classification tasks on two datasets which are consist of  long-form videos. \textbf{(1) COIN}~\cite{tang2019coin} is a large-scale dataset for comprehensive instructional video analysis. It consists of 11,827 videos related to 180 different tasks in 12 domains (e.g., vehicles, gadgets, etc.) related to our daily life. We use the official split and report the result on the test set.
\textbf{(2) LVU}~\cite{wu2021towards-longform} is a benchmark that contains 9 tasks
for evaluating long-form video understanding. It contains \textasciitilde30K videos from \textasciitilde3K movies from MovieClips \footnote{We download the videos from \url{https://www.movieclips.com}. \textasciitilde 90\% videos can be downloaded while others are broken.}. We choose content understanding (relationship, speaking style, scene/place) tasks for evaluation. We use the official split and report the result on test set.

\textbf{Implementation Details.}
We follow the setting of previous video-language pre-training models~\cite{lei2021clipbert,miech2020milnce,sun2019videobert,xu2021videoclip} used for video classification tasks. We only use the video encoder with a linear layer on the top and fine-tune the model for video classification with a cross-entropy loss. We fine-tune the model with 8 NVIDIA Tesla V100 GPUs. 

\begin{figure}[t]
    \centering
    \includegraphics[width=\linewidth]{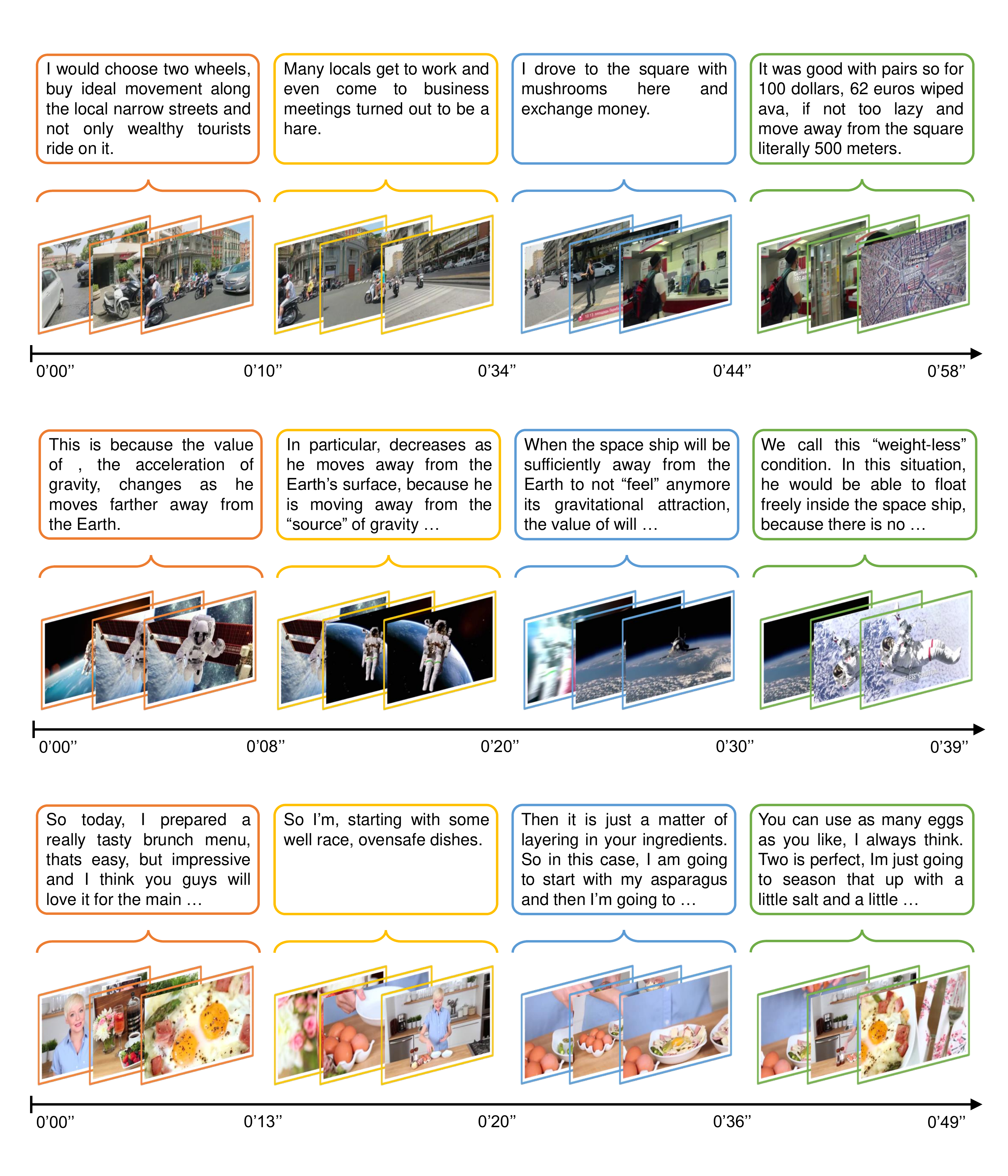}
    \caption{More examples of LF-VILA-8M.}
    \label{fig:lfvila_case}
\end{figure}

\section{LF-VILA-8M Dataset Details}\label{sec:app_dataset}

\begin{figure}[t]
    \centering
    \subfloat[Distribution of video duration.]
    {\begin{minipage}[b]{0.32\textwidth}
        \centering
        \includegraphics[width=\linewidth]{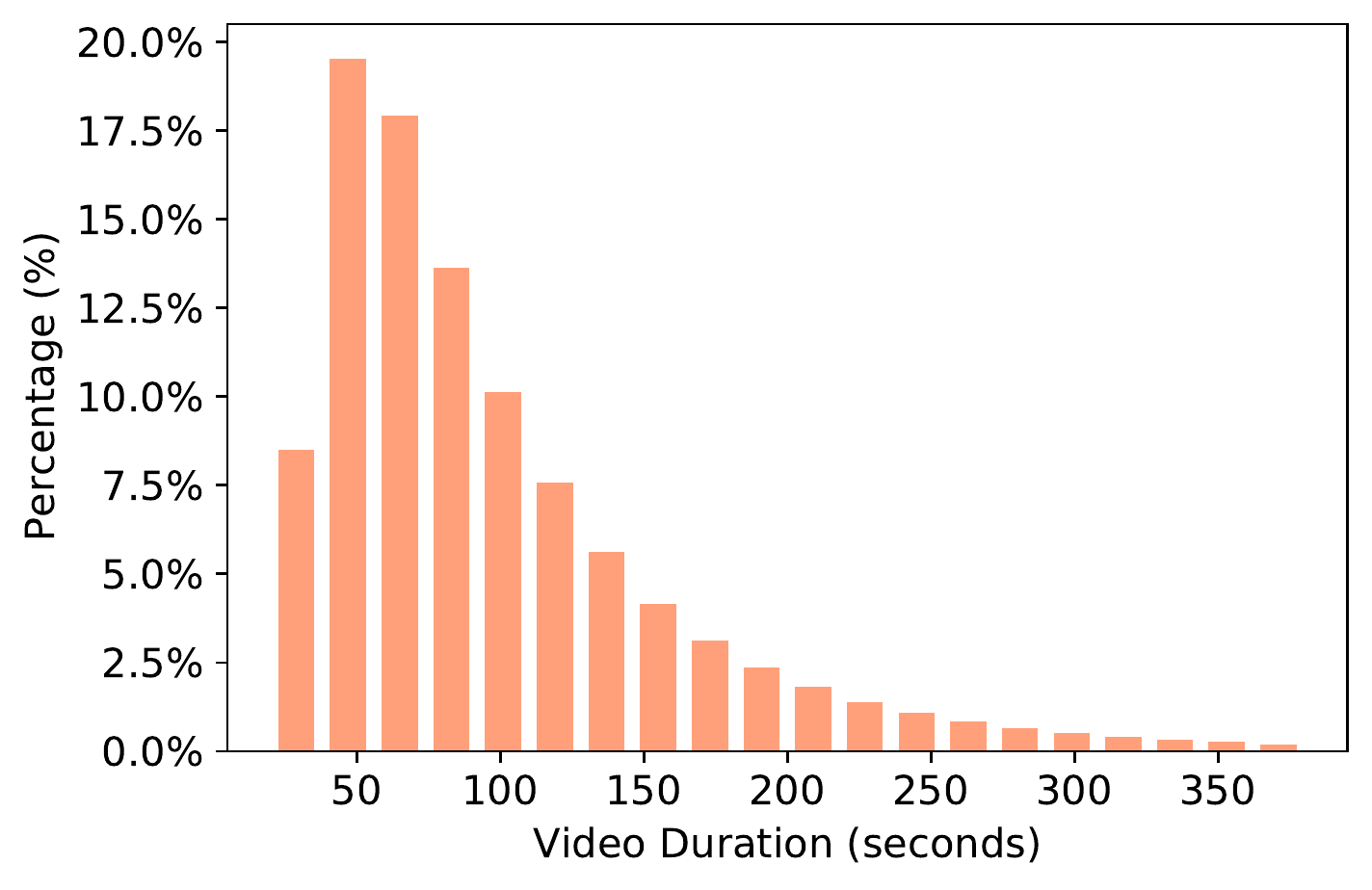}
        \label{fig:video_dur}
    \end{minipage}}
    \hfill
    \subfloat[Distribution of text length.]{
    \begin{minipage}[b]{0.32\textwidth}
        \centering
        \includegraphics[width=\linewidth]{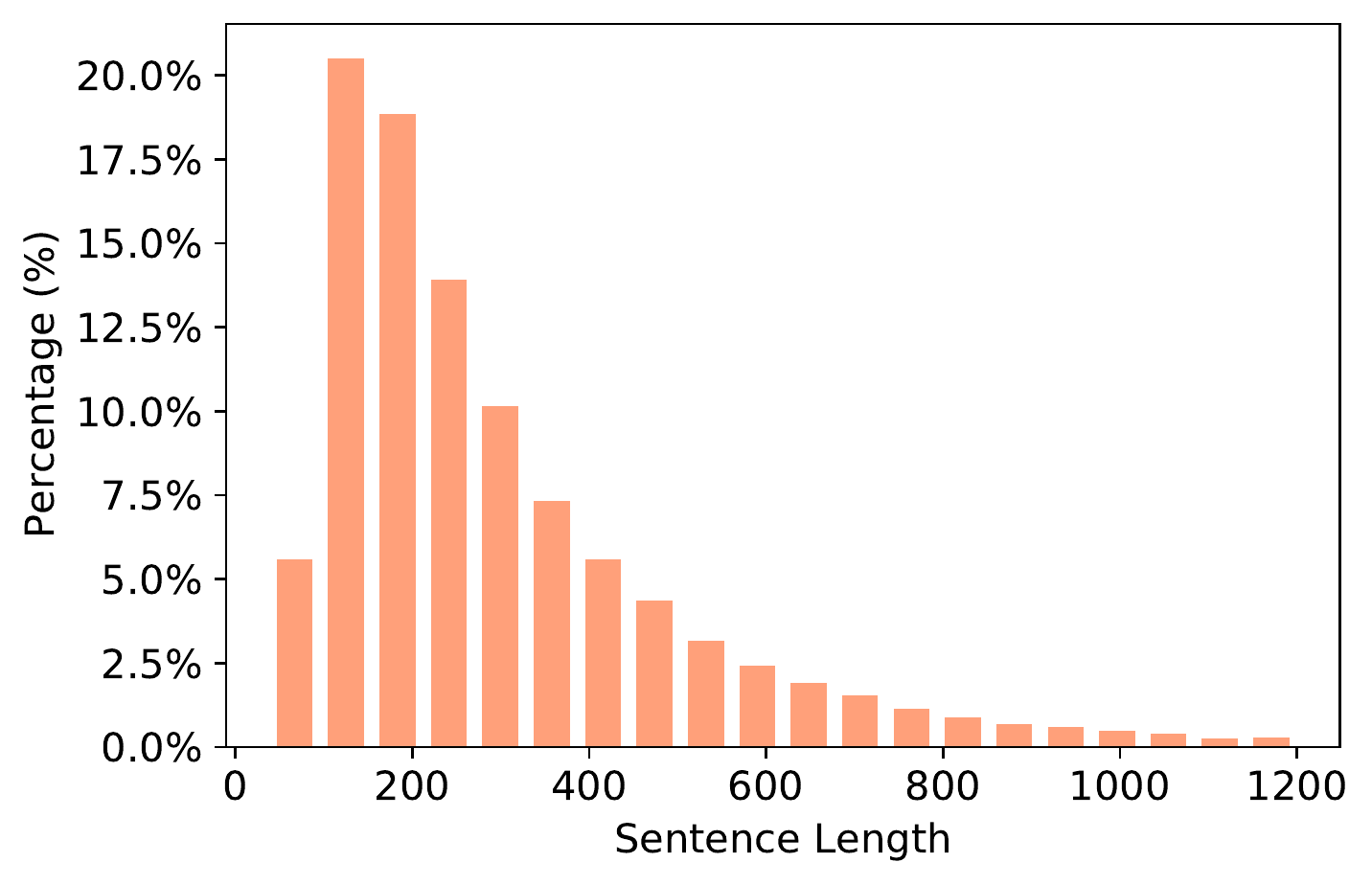}
        \label{fig:num_words}
    \end{minipage}}
    \hfill
    \subfloat[Distribution of the number of clip-sentence pairs.]{
    \begin{minipage}[b]{0.32\textwidth}
        \centering
        \includegraphics[width=\linewidth]{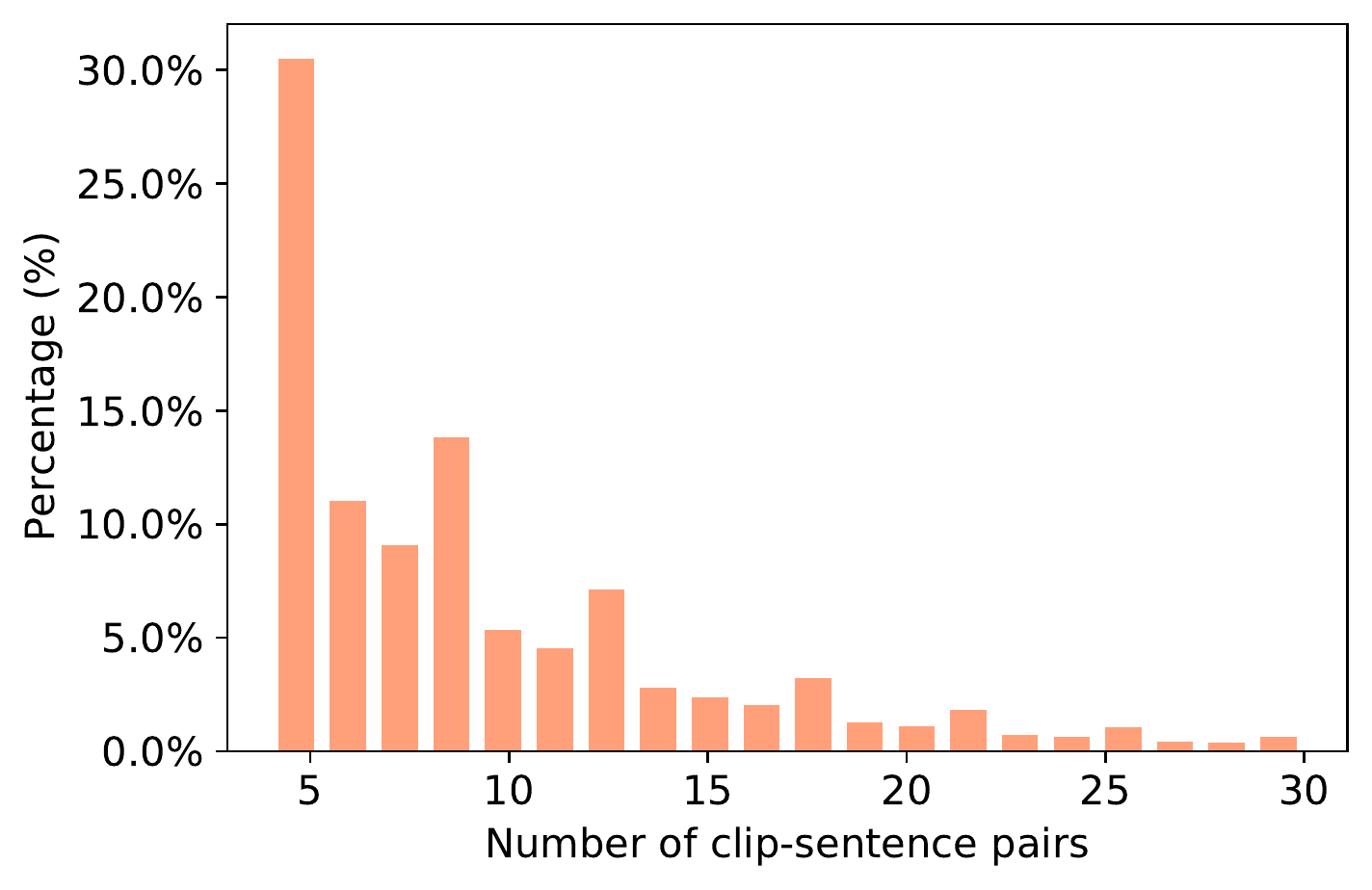}
        \label{fig:num_clips}
    \end{minipage}}
\caption{More detailed statistics of LF-VILA-8M dataset.}
\label{fig:lfvila_statistics}
\end{figure}

LF-VILA-8M is the largest long-form video-language dataset.
We show more examples of long-form video-paragraph pairs in Fig.~\ref{fig:lfvila_case}.
In Fig.~\ref{fig:lfvila_statistics}, we plot the histogram of video duration, text length, and the number of clip-sentence pairs.

\section{Limitation and Broader Impact.}\label{sec:limitation}
This paper has a broader impact on many video-language understanding applications such as video-text retrieval, video question-answering, etc. Since we apply two-stream architecture in the first stage, we can also utilize the single-modality features (i.e., video and language) from our model for even broader tasks. By learning vision-language representation from unlabeled videos and subtitles, our work may be easily extended and scaled to larger data.
On the other hand, vision-language pre-training may learn biased or offensive content from user-generated video-subtitle data. This may cause an improper understanding of videos. However, these concerns are general to the entire field and are not amplified by this work.

\section{Responsible AI Considerations} \label{sec:rai}
The proposed video-language dataset and pre-training model shows the capacity and generalization of learned VL representation which could benefit many applications of CV and NLP with a large range of uses across many domains. Each one of the uses has potential benefits and societal impacts. While we foresee that our technology could be used to find key information and improve efficiency and effectiveness for helpdesks, recommendation, retail and sales, we realize that it could also be used, in combination with new data, to fine-tune models to mislead, or otherwise harm people. We are also aware that this work uses considerable computation resources which itself, has environmental impacts. Therefore reducing the model size and computing effort is essential for future research.

Machine learning systems can display unfair behavior for different individuals or groups. This is a multi-dimensional, socio-technical challenge and is not explicitly addressed or captured in the current accuracy metrics for this research technology. In general, standardized fairness measures have not yet been agreed upon in academia or industry. We see opportunities for more work in this area to develop methods and benchmarks for measuring fairness aspects.

Given that user generated data is used, it is possible that certain demographic groups may not have enough representation. While we have balanced various video categories to mitigate for disparities, it is still likely that bias and fairness issues exist; this is an area of potential future work.  There may be a Western heteronormative bias, stereotypical depictions of historically marginalized populations and/or lack of representation among some groups. Although we have filtered the input data for explicit and violent content, it is possible that it hasn’t been totally eliminated in the training data and could have impacts on the results.

While some mitigations for potential harms can be done in the base model, it’s important to recognize that considering risks for fine-tuning data for particular scenarios is critical as well. Ultimately, choosing the application scenario of any final model used in a production system will require careful consideration of potential harms specific to the scenario.

\end{document}